%% file: main.tex
\documentclass[letterpaper, 10 pt, conference]{ieeeconf}  

\IEEEoverridecommandlockouts                              

\usepackage{array}
\usepackage{makecell}
\usepackage{hhline}
\usepackage{colortbl}
\usepackage{graphicx}
\usepackage{mathptmx} 
\usepackage{times} 
\usepackage{amsmath} 
\usepackage{amssymb}  
\usepackage{graphicx}
\usepackage{capt-of}
\usepackage{booktabs}
\usepackage{cuted}
\usepackage{booktabs}

\makeatletter
\let\NAT@parse\undefined
\makeatother

\usepackage{xcolor}
\usepackage{hyperref}
\definecolor{lightblue}{rgb}{0.68, 0.85, 0.9}
\definecolor{darkblue}{rgb}{0.48, 0.65, 0.7}
\definecolor{lightgray}{gray}{0.9}
\definecolor{citecolor}{HTML}{0071BC}
\hypersetup{colorlinks,linkcolor={red},citecolor={citecolor}}

\title{\LARGE \bf
SldprtNet: A Large-Scale Multimodal Dataset for CAD Generation in Language-Driven 3D Design
}

\author{Ruogu Li$^{\dagger}$ Sikai Li$^{\dagger}$ Yao Mu$^{\ddagger}$ and Mingyu Ding$^{\dagger}$
\thanks{$^{\dagger}$ University of North Carolina at Chapel Hill $^{\ddagger}$ Shanghai Jiao Tong University}
\thanks{This work has been accepted for publication in Proceedings of the IEEE International Conference on Robotics and Automation (ICRA), 2026}
}

\begin{document}

\maketitle
\thispagestyle{empty}
\pagestyle{empty}




\begin{abstract}

We introduce SldprtNet, a large-scale dataset comprising over 242,000 industrial parts, designed for semantic-driven CAD modeling, geometric deep learning, and the training/fine-tuning of multimodal models for 3D design. The dataset provides 3D models in both .step and .sldprt formats to support diverse training and testing. To enable parametric modeling and facilitate dataset scalability, we developed supporting tools, an encoder and a decoder, which support 13 types of CAD commands and enable lossless transformation between 3D models and a structured text representation. Additionally, each sample is paired with a composite image created by merging seven rendered views from different viewpoints of the 3D model, effectively reducing input token length and accelerating inference. By combining this image with the parameterized text output from the encoder, we employ the lightweight multi-modal language model Qwen2.5-VL-7B to generate a natural language description of each part’s appearance and functionality. To ensure accuracy, we manually verified and aligned the generated descriptions, rendered images, and 3D models. These descriptions, along with the parameterized modeling scripts, rendered images, and 3D model files, are fully aligned to construct SldprtNet. To assess its effectiveness, we fine-tuned baseline models on a dataset subset, comparing image-plus-text inputs with text-only inputs. Results confirm the necessity and value of multi-modal datasets for CAD generation. It features carefully selected real-world industrial parts, supporting tools for scalable dataset expansion, diverse modalities, and ensured diversity in model complexity and geometric features, making it a comprehensive multimodal dataset built for semantic-driven CAD modeling and cross-modal learning.

\end{abstract}

\input{sections/Introduction}
\input{sections/Related_work}
\input{sections/Dataset}

\input{sections/Supported_application}

\input{sections/Conclusion}






\bibliographystyle{IEEEtran}
\bibliography{main}

\end{document}

%% file: sections/Introduction.tex
\input{tables/dataset_comparison}
\section{INTRODUCTION}
Computer-Aided Design (CAD) plays a critical role in mechanical design and manufacturing, offering significant advantages over traditional paper-based drafting. Unlike manual drawings, CAD allows intuitive visualization of a part’s shape and dimensions and simplifies modifications. SolidWorks, a powerful CAD platform, has become a default choice for many mechanical designers. The .sldprt format, SolidWorks’ native part file, records the feature operations and parameters used in model creation, enabling quick iteration and flexible editing of designs. This parametric, feature-based representation ensures higher precision and editability than discrete 3D formats like point clouds or meshes.
\begin{figure}[t]
\centering
\includegraphics[width=\columnwidth]{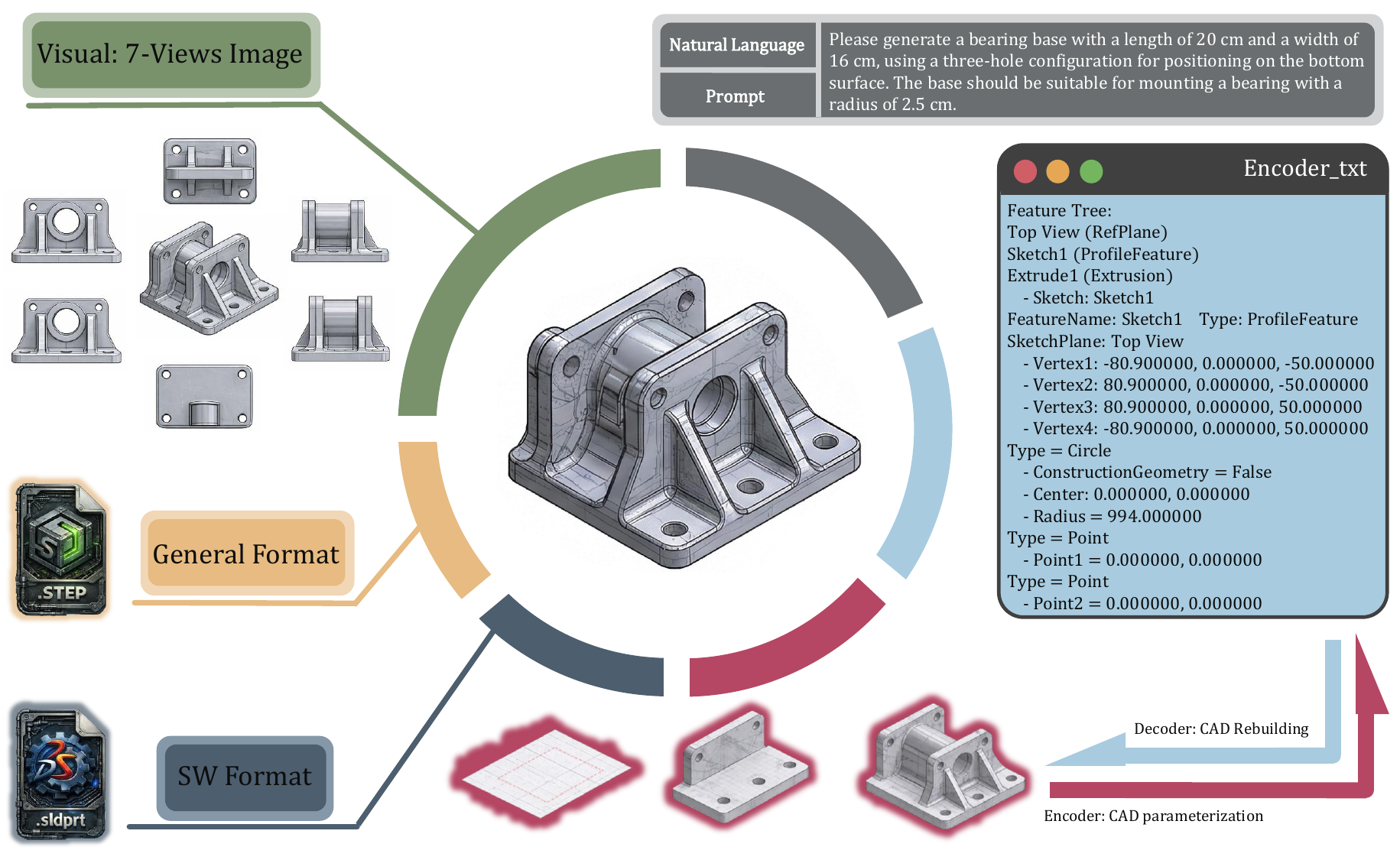}
\caption{Overview of SldprtNet dataset.}
\label{fig:teaser}
\end{figure}
Compared to other categories of 3D model datasets, CAD datasets are unique in that each sample must be manually created using specialized software. The high skill and time required for quality CAD modeling result in datasets that are much smaller in scale than image or text datasets. Furthermore, limited data quantity and quality, annotation difficulties, and the lack of a standardized parametric representation format for 3D models have constrained progress in this area. As a result, despite the surge of interest in LLMs, research on semantic-driven CAD modeling tasks remains in its early stages. In this study, our main contributions are summarized as follows:\\
\noindent $\bullet$ \textbf{Dataset:} We construct a large-scale CAD dataset of over 242,000 high-quality samples. Each sample contains a multi-view composite image, a structured parametric text representation, and a natural language description of the part’s appearance and functionality. SldprtNet is specifically designed to support semantic-driven CAD modeling tasks and geometric deep learning applications.\\
\noindent $\bullet$ \textbf{Parametrization Tools:} Leveraging the SolidWorks API, we develop two tools, encoder and decoder (detailed in Section 4), to convert .sldprt files into parameterized text and to reconstruct 3D parts from these text scripts. These tools enable lossless transformation between CAD models and text, effectively bridging the gap between natural language and CAD modeling. Unlike prior works limited to simple sketch-and-extrude sequences, our tools support 13 types of CAD operations, greatly expanding the diversity of 3D models in the dataset.\\
\noindent $\bullet$ \textbf{Multimodal Data Alignment:} For each 3D model, we provide a composite image (merging six orthographic views and one isometric view) and the corresponding parametric modeling script (Encoder\_txt). Using these as input, a multimodal language model (Qwen2.5-VL-7B) generates a natural language description (Des\_txt) that captures the part’s visual appearance and functional semantics. This yields a fully aligned set of images, parametric scripts, and descriptions for every sample. Such a multimodal dataset offers rich supervision signals and a strong foundation for advancing semantic-driven CAD modeling research.

%% file: tables/dataset_comparison.tex
\begin{table*}[t]
\centering
\caption{Comparative Overview of Major CAD and 3D Model Datasets}
\label{tab:dataset_comparison}
\renewcommand{\arraystretch}{1.2}

\begin{tabular*}{\textwidth}{@{\extracolsep{\fill}}lcccccc}

\specialrule{1.5pt}{0pt}{0pt}   

\textbf{Dataset} & \textbf{SldprtNet} & \textbf{ABC} & \textbf{ShapeNet} & \textbf{ModelNet} & \textbf{Fusion 360} & \textbf{Thingi10K} \\

\specialrule{0.6pt}{0pt}{0pt}   

\textbf{Models} & 242,606 & 1,000,000+ & 3,000,000+ & 151,128 & 50,000+ & 10,000 \\
\textbf{Format} & Sldprt & B-Rep & Mesh & Mesh & B-Rep & Mesh \\
\textbf{Parametric} & \checkmark & \checkmark & $\times$ & $\times$ & \checkmark & $\times$ \\
\textbf{Multi-view} & \checkmark & $\times$ & \checkmark & $\times$ & $\times$ & $\times$ \\
\textbf{Reconstructable} & \checkmark & $\times$ & $\times$ & $\times$ & $\times$ & $\times$ \\
\textbf{Model Description} & \checkmark & $\times$ & $\times$ & $\times$ & $\times$ & $\times$ \\

\specialrule{1.5pt}{0pt}{0pt} 

\end{tabular*}
\end{table*}

%% file: sections/Related_work.tex
\section{RELATED WORK}
Text-to-CAD modeling lies at the intersection of 3D geometry processing, CAD, and natural language understanding. While recent advances in mesh-based shape generation and vision-language pre-training have made notable progress, semantic-driven datasets and models specifically designed for CAD modeling remain limited. We briefly review mainstream 3D model datasets and recent work on language-guided CAD generation, and analyze their limitations. A comparative summary of representative datasets is provided in Table~\ref{tab:dataset_comparison}.\\
\noindent $\bullet$ \textbf{Non-Parametric 3D Datasets:} Early large-scale datasets such as ModelNet \cite{Wu2015ShapeNets} and ShapeNet \cite{Chang2015ShapeNet} are widely adopted in vision and graphics research. They represent 3D geometry using meshes or point clouds, capturing only final surface shapes without retaining any feature-based modeling information. For example, ModelNet includes ~150,000 models across various object categories, while ShapeNet provides ~3 million with rich class labels and annotations. Thingi10K \cite{Zhou2016Thingi10K} offers 10,000 printable models with diverse styles. Point-cloud-specific datasets such as ModelNet40, ScanNet \cite{Dai2017ScanNet}, SemanticKITTI \cite{Behley2019SemanticKITTI}, and ShapeNetPart have become benchmarks for perception \cite{Yi2017SyncSpecCNN}, reconstruction, and segmentation. However, none of these—whether mesh or point cloud—preserve design history, parametric operations, or sketch-based features. As a result, they lack parametric re-editability and cannot support semantically meaningful modifications. For instance, dimensions or constraints in these formats cannot be adjusted directly. While effective for classification, segmentation, and rendering, these representations are inherently limited for CAD automation, parameterized modeling, and language-guided design reasoning.\\
\noindent $\bullet$ \textbf{Parametric CAD Datasets:} To address the limitations of purely geometric datasets, several parametric CAD datasets have emerged. The ABC dataset contains over one million B-rep CAD models with precise surface and boundary representations, making it valuable for geometric learning \cite{Koch2019ABC}. However, it lacks modeling sequences and any textual annotations, which limits its use for sequential or language-driven generation tasks. The Fusion 360 Gallery dataset \cite{Willis2021Fusion360Gallery} includes real-world CAD modeling histories (e.g., sketches and extrusions), offering step-by-step modeling sequences. Nevertheless, it focuses primarily on basic extrusion-based models and lacks semantic-level textual annotations. Thus, while these datasets improve geometric precision, they remain insufficient for tasks requiring semantic understanding or language-based design generation.\\
\begin{figure*}[t]
    \centering
    \includegraphics[width=\textwidth]{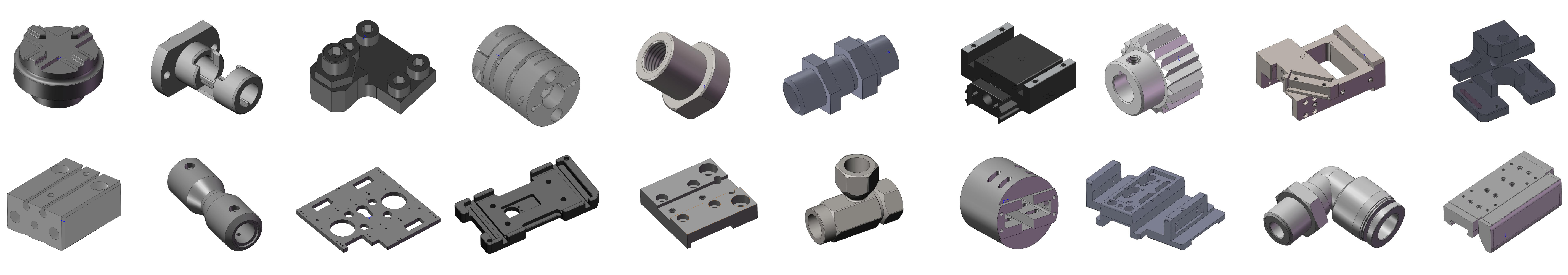}
    \caption{Twenty randomly sampled CAD parts from SldprtNet, illustrating its industrial orientation and geometric diversity across varied mechanical components.}
    \label{fig:model_samples}
\end{figure*}
\noindent $\bullet$ \textbf{Language-Guided CAD Generation:} Recent efforts have explored CAD generative modeling using language or sequential modeling approaches. DeepCAD introduced a transformer-based network to model sequences of CAD commands \cite{Wu2021DeepCAD}. It was trained on ~170,000 CAD sequences (a subset of ABC with modeling histories) and pioneered framing CAD modeling as a language modeling problem. However, DeepCAD was limited to only two command types (2D sketching and extrusion), lacked support for complex feature workflows, and did not incorporate natural language input. To bridge this gap, Text2CAD proposed generating CAD models guided by textual prompts \cite{Khan2024Text2CAD}. It augmented Deep-CAD’s data with synthetic language descriptions (ranging from beginner to expert level) to form paired examples of CAD sequences and text. While this established a link between CAD operations and natural language, the synthetic nature of the text can lead to semantic misalignment with the actual geometry. Moreover, the absence of any image modality in Text2CAD’s data restricts perceptual grounding and limits generalization.\\
\noindent $\bullet$ \textbf{Toward Multimodal CAD Learning:} Most existing CAD datasets focus on a single modality—geometry only (e.g., ABC), modeling sequence only (e.g., DeepCAD), or text prompts only (e.g., Text2CAD). In contrast, modern multimodal models like CLIP \cite{Radford2021CLIP}, Flamingo \cite{Alayrac2022Flamingo}, and BLIP-2 \cite{Li2023BLIP2} have shown that aligned cross-modal learning is key to generalization, cross-task transfer, and zero-shot reasoning. In the CAD domain, a multimodal dataset that combines precise geometry (3D CAD models), rendered images (multi-view projections), structured modeling sequences (parametric CAD commands), and natural language descriptions can significantly enhance a model’s capability in understanding and generation. For example, Flamingo demonstrates strong few-shot performance on vision-language tasks by leveraging cross-modal alignment during training, while BLIP-2 shows that even a limited amount of paired data can be leveraged effectively via modality bridging. Therefore, for the Text-to-CAD task, constructing a unified multimodal dataset that integrates images, natural language, 3D geometry, and modeling sequences not only provides richer supervision but also lays a foundation for general-purpose pretraining, improving downstream performance and transferability. \\
Beyond the aforementioned datasets, several recent works have attempted to extend large language models toward CAD generation, such as CAD-GPT \cite{Xu2024CADMLLM}, CAD-MLLM \cite{Sanghi2022EditNet}, CAD-Coder \cite{Wang2025CADGPT}, and CAD-Llama \cite{Doris2025CADCoder}. While these studies demonstrate the potential of multimodal or code-driven approaches, they still exhibit notable limitations in key aspects. CAD-GPT emphasizes the spatial reasoning capabilities of multimodal LLMs, but its data is only a small-scale extension of DeepCAD, restricted to sketching and extrusion commands, and lacks large-scale industrial coverage. CAD-MLLM introduces the modality-rich Omni-CAD dataset, yet a significant portion of its samples are synthetically generated rather than real engineering parts, and it does not provide executable modeling sequences for full reconstruction. CAD-Coder is the first open-source vision–language model capable of translating images into CadQuery code, but its reliance on the lightweight CadQuery DSL deviates from industrial CAD workflows, and its dataset size (163k) and command coverage remain limited. CAD-Llama proposes a structured parametric CAD code (SPCC) and hierarchical annotation, but it is primarily evaluated on synthetic or small-scale subsets, and lacks multimodal alignment such as rendered views or natural language paired with real-world parts.

%% file: sections/Dataset.tex
\section{DATASET}
\subsection{Principles and Requirements}
Given the challenges in Text-to-CAD modeling and the growing demands of multimodal generative tasks, we revisit and extend the classical dataset criteria proposed in the ABC dataset \cite{Hunde2022FutureProspectsCAD}. While traditional benchmarks emphasize large scale, geometric fidelity, parametric representation, and category balance, they often fall short in supporting language-driven, editable, and multimodal modeling workflows \cite{Dupont2022CADOpsNet}. To bridge this gap, we define five expanded requirements for next-generation CAD datasets:\\
\noindent $\bullet$ \textbf{Multimodality:} CAD modeling inherently involves geometry, procedural logic, visual perception, and natural language \cite{Lambourne2023CLIPSculptor}. However, most existing datasets provide only geometry or command sequences. For instance, ABC dataset lacks language or visual content, while Text2CAD omits rendered images. An ideal dataset should integrate 3D models, multi-view images, structured modeling instructions, and natural language descriptions to enable joint learning across modalities and support tasks such as text-guided modeling or visual-to-CAD generation.\\
\noindent $\bullet$ \textbf{Bidirectional Representability:} Unlike static output tasks in vision or NLP, CAD modeling emphasizes procedural traceability and logical consistency. An ideal dataset should support both forward modeling, which converts structured instructions into 3D geometry, and reverse modeling, which derives modeling commands from geometry or images. This bidirectional design enables closed-loop training, error detection, synthetic data generation, and controllable model editing \cite{Para2021SketchGen}.\\
\noindent $\bullet$ \textbf{Semantic Annotation:} Unlike image datasets with clear class boundaries, CAD parts are shaped by engineering intent, which makes function-level semantics more meaningful than conventional category labels. Annotating parts with functional roles, such as shaft, support, or guide rail, facilitates tasks such as part retrieval, function-aware generation, and abstraction. \cite{Manda2021CADSketchNet}, \cite{Chen2021TransUNet}.\\
\noindent $\bullet$ \textbf{Editability:} Static mesh datasets do not support structured editing. CAD datasets should retain parametric information and feature hierarchies to enable flexible modifications through standard tools or code-based interfaces. This supports use cases like language-driven editing, prompt tuning, and iterative design refinement.\\
\noindent $\bullet$ \textbf{Human-Readable Format:} Modeling scripts should use interpretable syntax, clear feature hierarchies, and consistent naming. This not only enhances debugging and annotation but also facilitates human-AI collaboration and alignment with design intent \cite{Yamamoto2005InteractionDesign}.
\subsection{Composition and Format}
We curated SldprtNet by collecting over 242,000 industrial CAD part models from publicly available online repositories, including GrabCAD, McMaster-Carr, and FreeCAD \cite{Machado2019ParametricCADModeling}. Each sample in the dataset comprises four components:\\
\noindent $\bullet$ \textbf{3D Models:} Each part is provided in two formats.\\
\begin{figure*}[t]
    \centering
\includegraphics[width=0.8\textwidth]{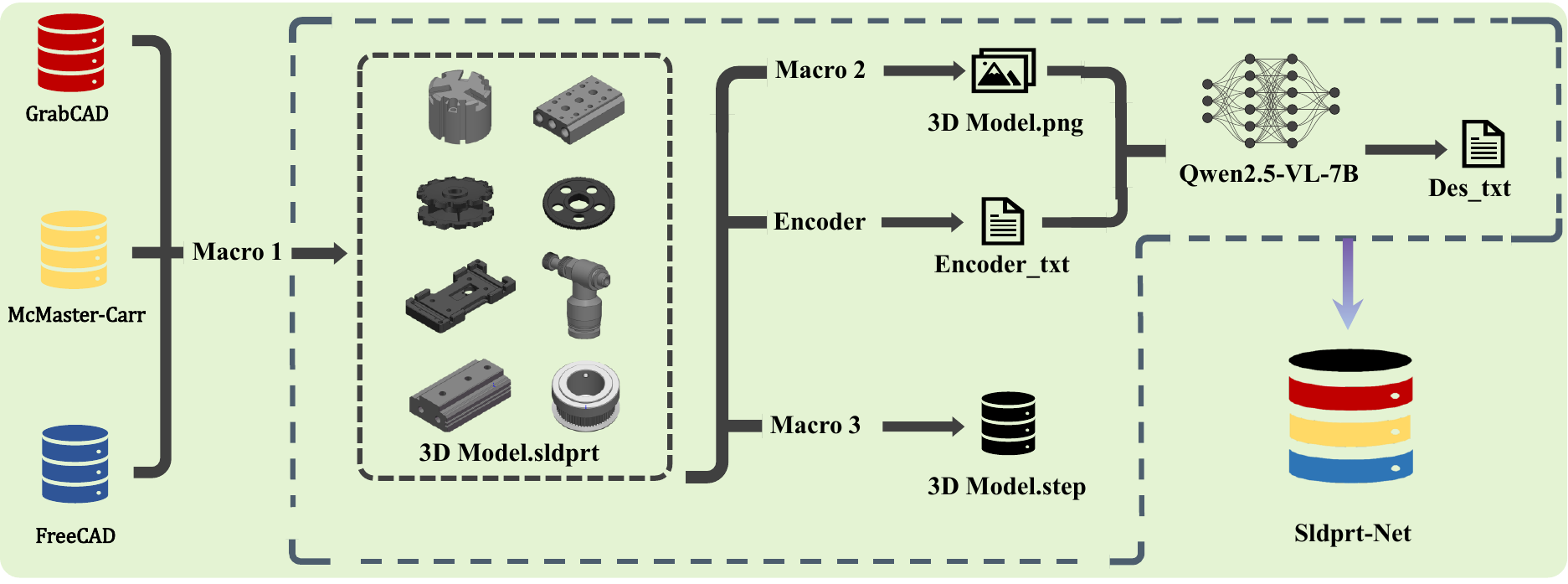}
    \caption{Overview of the automated data processing pipeline for constructing SldprtNet.}
    \label{fig:pipeline}
\end{figure*}
(1) sldprt the native SolidWorks part file, obtained via manual modeling. It encodes the feature-tree history (commands, parameters, and modeling sequence) and allows lossless reconstruction and feature-level semantic extraction for our purposes.\\
(2) step – a neutral, standardized 3D CAD exchange format that supports cross-platform interoperability. We include this for validation and use with other CAD software.\\
\noindent $\bullet$ \textbf{Images:} For each CAD model, we render seven PNG images from different standard views: six orthographic (front, back, left, right, top, bottom) and one isometric. These multi-view images comprehensively capture the 3D geometry in 2D form and serve as visual inputs for vision-language models. To optimize token usage during multimodal model inference, we composite all seven views into a single image. This reduces input length and accelerates processing without compromising visual completeness.\\
\noindent $\bullet$ \textbf{Parametric Text:} Using our custom Encoder tool (Section 4), each .sldprt file is converted into a structured text script encoding the CAD modeling instructions. The output (Encoder\_txt) is a human- and machine-readable description of the model’s construction. Table~\ref{tab:parametric_text} shows an excerpt of this format, including (1) a Feature Tree listing all features used (e.g., Extrude, Fillet) in modeling order with their names and parent-child relationships, and (2) detailed parameters for each feature (dimensions, constraints, sketch entities, dependencies). This structured format exposes the modeling logic explicitly and facilitates learning of CAD sequences by generative models \cite{Cherenkova2020PvDeconv}, \cite{Hong2023AlignYourGaussians}.\\
\noindent $\bullet$ \textbf{Natural Language Descriptions:} Unlike prior works (e.g., Text2CAD) that rely on modeling sequences or sketches alone to generate text, our dataset includes full visual information for each part \cite{Jain2021DreamFields}. We employ a multimodal LLMs, Qwen2.5-VL-7B, providing it both the composite rendered image and the parametric text as input to produce a descriptive caption for the part \cite{Jain2021DreamFields}. The inference process was run on 12 NVIDIA A100 GPUs for 368 GPU-hours to generate more than 242,000 parts descriptions, each capturing both functional and geometric aspects of the design. Qwen2.5-VL-7B uses a vision transformer to integrate visual features with the modeling text, enabling fine-grained alignment between the geometry and the description \cite{Poole2022DreamFusion}, \cite{Liu2021SwinTransformer}. This allows the model to pick up details like hole patterns, contour outlines, and aspect ratios in context with the part’s functional intent \cite{Mildenhall2021NeRF}. The resulting descriptions are structurally consistent with the CAD modeling steps, clearly convey the part’s functionality and features, and remain well-aligned with the visual geometry of the model \cite{He2023T3Bench}.
\subsection{Processing Pipeline}
To construct the multimodal SldprtNet dataset encompassing graphical, parametric, and textual modalities, we designed a fully automated pipeline for data acquisition and annotation in Fig.~\ref{fig:pipeline}. First, we collected approximately 680,000 industrial part models in .sldprt format from the three public CAD platforms mentioned above using an automated script (Macro 1). We then filtered these to ensure modeling completeness and diversity: we retained over 242,000 models that contain at least one of 13 representative feature types (the supported feature list is detailed in Section 4) \cite{Mallis2023SHARPChallenge}. For each remaining model, we performed the following steps within SolidWorks:\\
(1) Image Rendering (Macro 2): We rendered six orthographic views (front, back, left, right, top, bottom) and one isometric view of the part, then merged them into a single composite image (PNG). Fig.~\ref{fig:7views} shows an example of this multi-view composite.\\
(2) Parametric Command Extraction (encoder): We invoked our command-line encoder tool to extract the modeling sequence from the .sldprt file. This produces the structured text file (Encoder\_txt) described above, which serves as input to downstream learning models.\\
(3) Standard Format Conversion (Macro 3): We converted the .sldprt file into the standard .step format for cross-platform geometry compatibility and validation.\\
To generate natural language descriptions aligned with both the appearance and function of the 3D model, we take the rendered 3D model image (3D Model.png) and the parametric modeling script (Encoder\_txt) as inputs and perform inference using the Qwen2.5-VL-7B multimodal LLMs. Within Qwen2.5-VL-7B, a vision encoder aligns visual features with the parameterized modeling scripts across modalities, thereby generating semantically accurate and geometrically consistent natural language descriptions (Des\_txt). As a result, each sample in SldprtNet contains five aligned modalities: the source model (.sldprt), the standard geometry model (.step), the rendered image (.png), the parametric modeling script (Encoder\_txt), and the natural language description (Des\_txt).\\
This pipeline not only ensures consistency across modalities but also enhances the extensibility and applicability of the dataset for a wide range of multimodal learning tasks.

\begin{figure}[t]
    \centering
    \includegraphics[width=\columnwidth]{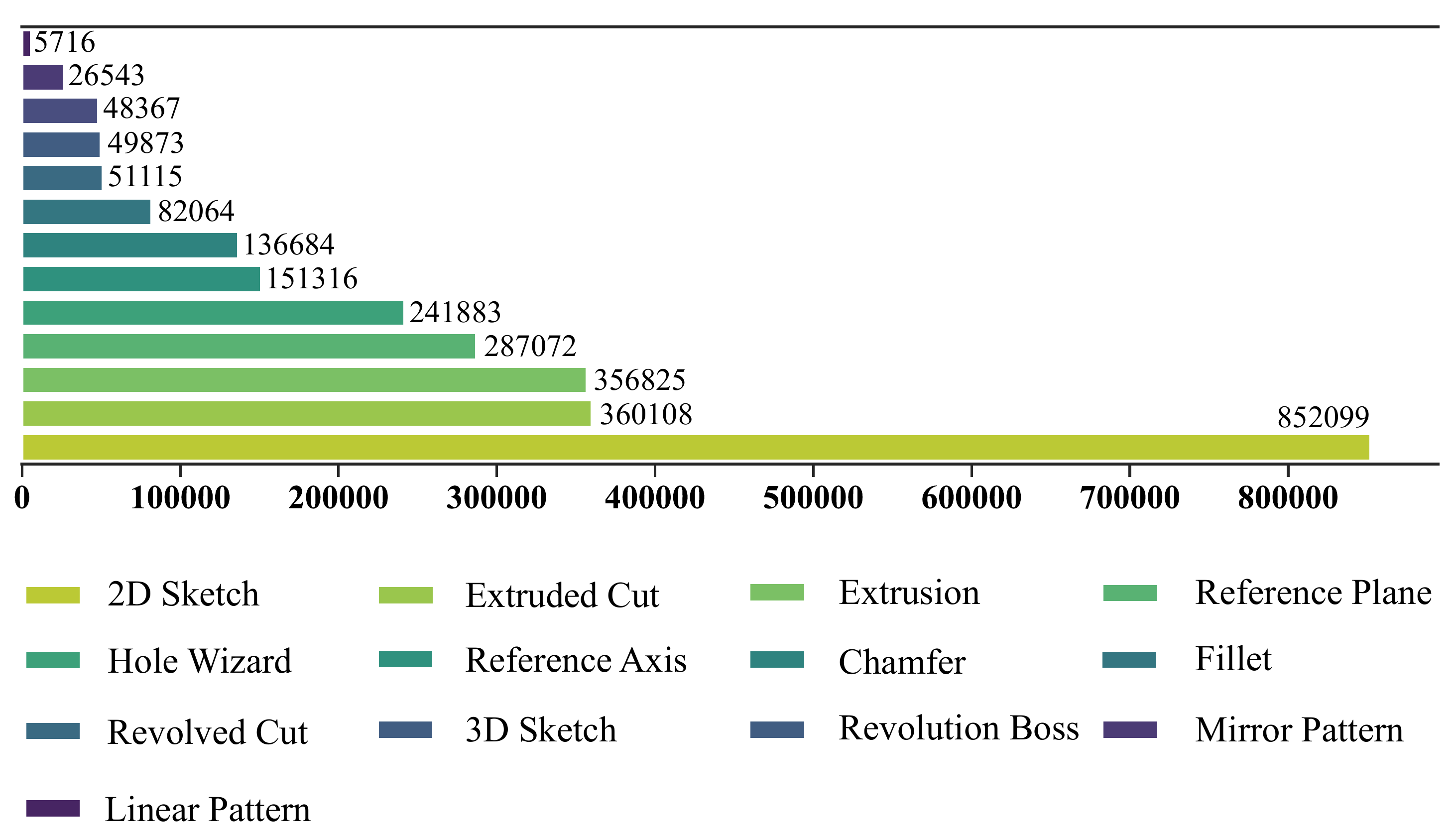}
    \caption{Distribution of CAD model complexity in the SldprtNet dataset based on the number of feature operations per part. The chart presents four complexity levels with their corresponding proportions and sample counts.}
    \label{fig:all_dataset_statistic}
\end{figure}

\subsection{Statistics and Analysis}
Following the completion of data acquisition and preprocessing, we analyzed the Feature Tree information from the Encoder\_txt files in the SldprtNet dataset \cite{Scarselli2009GNN}. Specifically, we extracted all the CAD feature types used across the 3D models, as well as their corresponding frequencies, to construct a comprehensive distribution chart, shown in Fig.~\ref{fig:all_dataset_statistic}.
The chart summarizes the usage statistics of 13 core CAD features, including 2D Sketch, Extrusion, Linear Pattern, among others. The key observations are as follows:\\
(1) 2D Sketch emerges as the most frequently used feature type, serving as the foundational step in nearly all models regardless of their complexity. This aligns with standard CAD workflows, where most 3D geometries originate from planar sketches.\\
(2) Chamfer and Fillet—commonly employed in mechanical component design for edge treatment and stress reduction—also exhibit high usage frequencies, suggesting that the dataset predominantly consists of industrial-grade parts.\\
(3) Linear Pattern and Mirror Pattern appear less frequently. These features are typically used for repetitive or symmetric geometries and can often be substituted through manual replication of base features. Their relatively low presence may also reflect a preference for explicit design over procedural patterning in the original models.

In addition to feature-type distribution, we conducted a complexity analysis of the dataset based on criteria adapted from DeepCAD \cite{Wu2021DeepCAD}, using the total number of modeling features (i.e., command operations) in each part as a proxy for its complexity. 
\begin{figure}[t]
    \centering
    \includegraphics[width=0.8\columnwidth]{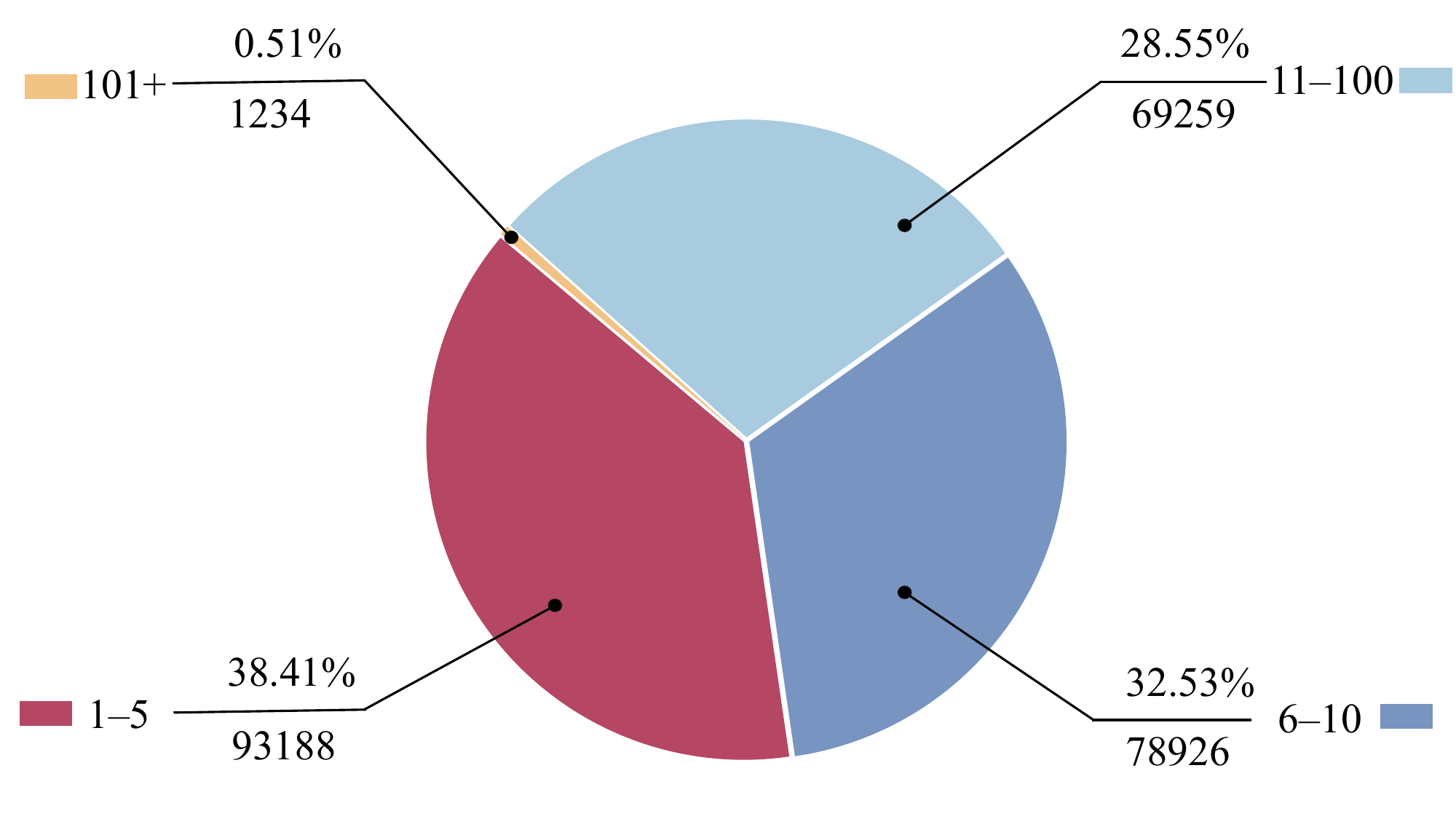}
    \caption{Distribution of CAD features and total counts in the SldprtNet dataset. The figure summarizes the usage frequency of 13 parametric operations across all parts.}
    \label{fig:feature_distribution}
\end{figure}
As shown in Fig.~\ref{fig:feature_distribution}, to quantify model complexity, we analyze the number of unique CAD commands recorded in the Feature Tree of each part’s Encoder\_txt file. This count reflects the number of sequential modeling steps involved in constructing the 3D geometry—higher values indicate more intricate design logic and geometric detail \cite{Shapiro1991CSGConstruction}. For instance, a model with a complexity score of 2 might involve two commands, such as Sketch and Extrude. Based on this definition, we categorize model complexity into four distinct levels:
\\
\noindent $\bullet$ \textbf{Level 1 (Simple):}  1–5 features, 93,188 samples.\\
\noindent $\bullet$ \textbf{Level 2 (Moderate):} 6–10 features, 78,926 samples.\\
\noindent $\bullet$ \textbf{Level 3 (Advanced):} 11–100 features, 69,259 samples.\\
\noindent $\bullet$ \textbf{Level 4 (Expert):} over 100 features, 1,234 samples.\\
Key insights from the complexity distribution include:\\
(1) The majority of the dataset is composed of simple, moderate, and advanced models in relatively balanced proportions. This diversity ensures robust training coverage across varying levels of modeling intricacy, making SldprtNet well-suited for multi-stage generative modeling tasks.\\
(2) Although expert-level models are rare, they represent highly intricate CAD workflows—often involving nested assemblies, complex constraints, and hybrid features. These models provide critical benchmarks for evaluating the reasoning depth and sequence modeling capacity of multimodal language models such as LLMs or vision-language transformers \cite{Chen2022PolyGen}.\\
(3) This hierarchical complexity taxonomy also supports stratified evaluation of downstream models. Given the wide variance in geometric logic and command length across complexity levels, performance can be better assessed through complexity-specific benchmarks.\\
(4) Moreover, the tiered classification offers a practical foundation for implementing curriculum learning. Initial model training may begin with simpler instances to bootstrap geometric comprehension, followed by gradual exposure to higher-complexity samples. This staged approach can improve convergence rates and model generalization, especially in long-horizon CAD sequence generation \cite{Wei2022ChainOfThought}, \cite{Williams1989RNNLearning}.\\
In summary, SldprtNet demonstrates a high degree of diversity in feature types and structural complexity, offering both broad coverage and deep granularity across the dataset.
\subsection{Base Line}
To evaluate the effectiveness of the multimodal dataset in CAD generation tasks, we fine-tuned two baseline models on a 50K-sample subset of the SldprtNet: Qwen2.5-7B (trained with Encoder\_text only) and Qwen2.5-7B-VL (trained with both image and Encoder\_text). As shown in Table~\ref{tab:baseline_comparison}, the multimodally trained Qwen2.5-7B-VL outperforms the text-only Qwen2.5-7B across several key metrics. Specifically, it achieves higher scores in Exact Match Score (0.0099 vs. 0.0058), Command-Level F1 (0.3670 vs. 0.3247), and Partial Match Rate (0.6162 vs. 0.5554), indicating a stronger alignment between the generated CAD commands and ground-truth modeling instructions. These results demonstrate that incorporating rendered geometric views as an auxiliary modality enhances the model’s understanding of geometric semantics and procedural logic.\\
\input{tables/data_type}
Although the text-only model slightly outperforms in Parameter Tolerance Accuracy (0.5016 vs. 0.4630), this may suggest a tendency toward overfitting on numeric values rather than capturing structural semantics. The baseline experiments are reproducible, and we will release all relevant code and dataset subsets through the anonymous repository referenced in the Code section. Overall, these findings validate the effectiveness of multimodal supervision in improving the performance of language-driven CAD generation models, and further underscore the importance of richly annotated datasets like SldprtNet for advancing research in generative CAD modeling.

%% file: tables/data_type.tex
\setlength{\arrayrulewidth}{0.6pt}
\renewcommand{\arraystretch}{1.05}
\setlength{\textfloatsep}{8pt}
\setlength{\floatsep}{6pt}
\setlength{\intextsep}{8pt}
\begin{table}[t]
\centering
\caption{Example of Parametric Text Representation}
\label{tab:parametric_text}
\vspace{0.6em}

\begin{tabular*}{\columnwidth}{@{\extracolsep{\fill}}p{\columnwidth}@{}}

\noalign{\vskip 4pt}
\noalign{\hrule height 1.2pt}
\noalign{\vskip 4pt}

\textbf{Feature Tree:} \\
Top View (RefPlane) \\
Sketch1 (ProfileFeature) \\
Extrude1 (Extrusion) \\
\quad - Sketch: Sketch1 \\

\hline

\textbf{FeatureName:} Top View \quad Type: RefPlane \\
\quad - Vertex1: -80.900000, 0.000000, -50.000000 \\
\quad - Vertex2: 80.900000, 0.000000, -50.000000 \\
\quad - Vertex3: 80.900000, 0.000000, 50.000000 \\
\quad - Vertex4: -80.900000, 0.000000, 50.000000 \\

\hline

\textbf{FeatureName:} Sketch1 \quad Type: ProfileFeature \\
SketchPlane: Top View \\
\quad - Vertex1: -80.900000, 0.000000, -50.000000 \\
\quad - Vertex2: 80.900000, 0.000000, -50.000000 \\
\quad - Vertex3: 80.900000, 0.000000, 50.000000 \\
\quad - Vertex4: -80.900000, 0.000000, 50.000000 \\

Type = Circle \\
\quad - ConstructionGeometry = False \\
\quad - Center: 0.000000, 0.000000 \\
\quad - Radius = 994.000000 \\

Type = Point \\
\quad - Point1 = 0.000000, 0.000000 \\

Type = Point \\
\quad - Point2 = 0.000000, 0.000000 \\

\hline

\textbf{FeatureName:} Extrude1 \quad Type: Extrusion \\
sd: True \\
flip: False \\
dir: False \\
t1: 6 \\
t2: 0 \\
d1: 150.000000 \\
d2: 0.000000 \\
dang1: 0 \\
dang2: 0 \\
merge: True \\
useFeatScope: True \\
useAutoselect: False \\
t0: 0 \\
startOffset: 0 \\
flipStartOffset: False \\

\noalign{\vskip 4pt}
\noalign{\hrule height 1.2pt}
\noalign{\vskip 4pt}

\end{tabular*}
\end{table}

%% file: sections/Supported_application.tex
\section{SUPPORTED APPLICATION}
This section focuses on two key tools developed during the construction of the SldprtNet dataset, namely the encoder and decoder, as well as the dataset’s applicability to the representative task of Text-to-CAD generation.\\
\subsection{Tools}
The encoder and decoder are two essential auxiliary tools developed as part of the dataset pipeline. Built upon the COM interface provided by SolidWorks and implemented via macro-based scripting, these tools enable automatic traversal and reconstruction of .sldprt part models. Together, they support 13 common CAD feature types and allow for a lossless, bidirectional transformation between parameterized CAD models and structured textual representations. Their implementations are as follows:\\
\noindent $\bullet$ \textbf{Encoder: CAD → Text:}\\
(1) Upon execution, the encoder automatically traverses the Feature Tree of an .sldprt file, extracting the feature types, feature names, and parent-child relationships in the order dictated by the modeling history \cite{Seff2020SketchGraphs}.
\input{tables/mutimodal_comparison}
An example is provided in the first five rows of Table~\ref{tab:parametric_text}.\\
(2) Based on the feature types retrieved from the Feature Tree, the encoder sequentially calls corresponding internal modules to extract detailed parameters for each feature. These parameter details are illustrated in Appendices Table~\ref{tab:parametric_text} following the initial rows.\\
(3) For each supported feature type, a consistent, human-readable, and machine-interpretable text format is adopted. After completing the above steps, the encoder generates an Encoder\_txt file, completing the parametric representation of the original .sldprt file.\\
\noindent $\bullet$ \textbf{Decoder: Text → CAD:}\\
(1) Upon execution, the decoder first initializes the modeling environment by creating a blank .sldprt part document.\\
(2) It then reads the encoder\_txt file and parses the entire Feature Tree section. Guided by the feature order and parent-child hierarchy, it sequentially accesses each feature’s parameters and invokes the SolidWorks API to create the corresponding feature within the modeling environment.\\
(3) Following the original modeling sequence, the decoder reconstructs each feature step-by-step, ensuring geometric and topological consistency with the source model.\\
Together, the encoder and decoder form a closed-loop modeling system, enabling the round-trip transformation from model to instruction and back to model. This capability provides structural verification, supports data augmentation, and opens avenues for the expansion of SldprtNet in future applications \cite{Mao2022ParametricModelAware}.

\begin{figure}[t]
    \centering
    \includegraphics[width=\columnwidth]{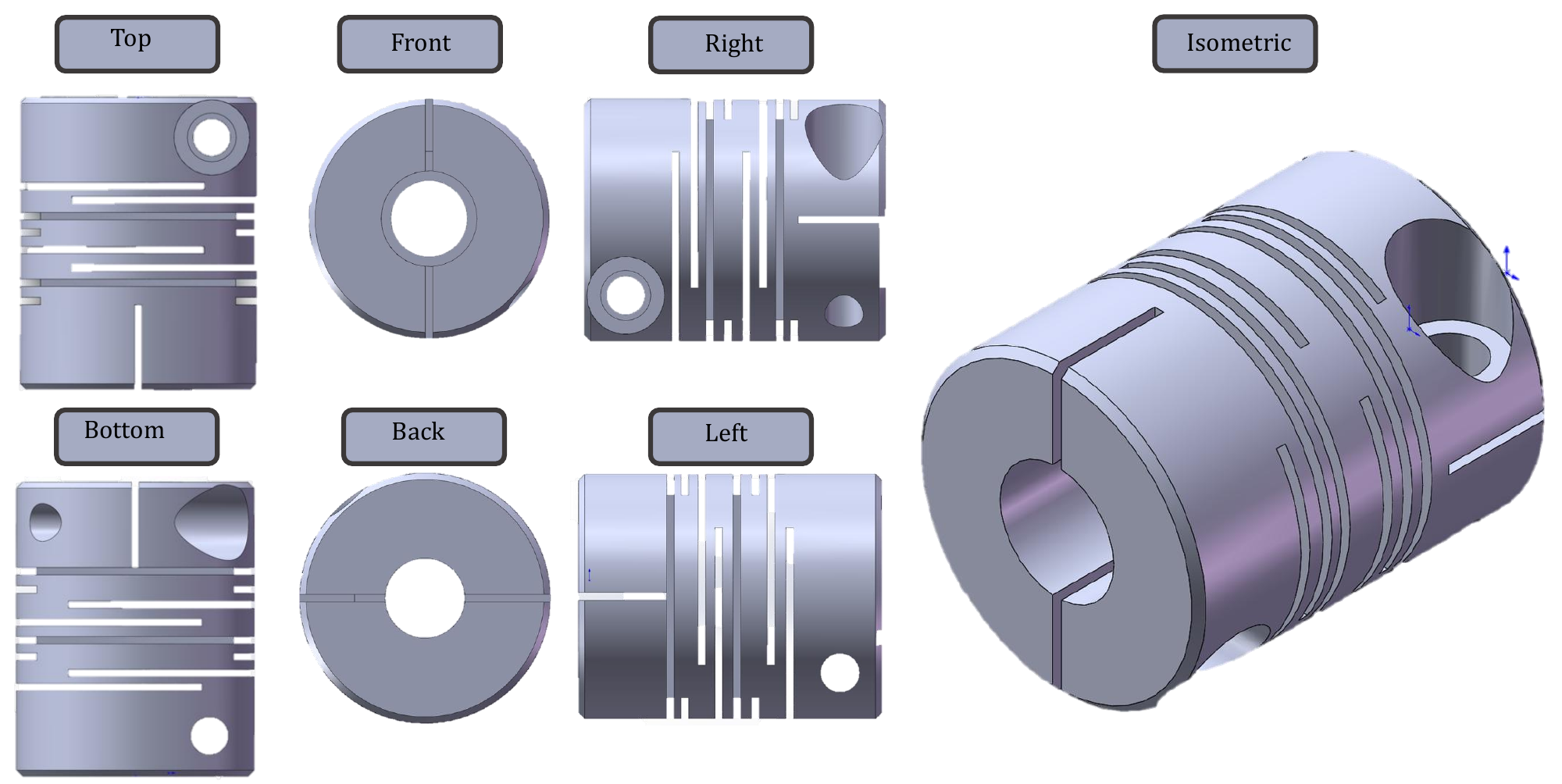}
    \caption{Composite image of a CAD model from seven standard views, including top, bottom, left, right, front, back, and isometric.}
    \label{fig:7views}
\end{figure}

\subsection{Capability in Supporting Text-to-CAD Tasks}
Based on its design and supporting tools, SldprtNet offers robust capabilities for Text-to-CAD tasks:\\
(1) Natural Language–Driven Parametric Modeling: The encoder converts CAD models into structured, semantically meaningful command sequences, mapping natural language descriptions to modeling operations \cite{Willis2021Fusion360Gallery}. During training, models learn the alignment between textual inputs and CAD procedures, and during inference, the language model maps natural language input (e.g., "Create a bracket with four threaded holes") into structured CAD commands \cite{Xu2023HierarchicalNeuralCoding}, executed by the decoder to generate the model. This "language → instruction → geometry" pipeline enhances the modeling process by capturing both intent and execution logic, significantly expanding the operational capacity of language in design \cite{Ritchie2023NeurosymbolicGraphics}.\\
(2) Multimodal Input and Cross-Modal Modeling: Each sample in SldprtNet contains rendered images, modeling instruction text, and natural language descriptions \cite{Seff2021Vitruvion}, making it a rich input source for vision-language models (VL-LLMs) \cite{Lin2023Magic3D}. This enables tasks such as image-text alignment, multi-view geometric reasoning, and cross-modal prompt learning.\\
(3) Controllable Modeling and Parameter-Level Editing: The encoder’s structured output provides semantic clarity, allowing the decoder to execute model-generated instructions directly \cite{Ganin2021CADAsLanguage}. This enables not only model creation from scratch but also fine-grained, controllable edits to existing models \cite{Mitra2013StructureAware}, such as adjusting dimensions or directions \cite{Du2018InverseCSG}.\\
(4) Complexity-Based Evaluation Mechanism: Models are categorized by the number of CAD features into complexity tiers (1–5, 6–10, 11–100, 101+), supporting curriculum-based training, benchmarking, and skill evaluation across Text-to-CAD tasks.\\
(5) Interpretability and Reverse Engineering Support: The closed-loop conversion between encoder and decoder enables semantic verification and geometric reconstruction of command sequences, ensuring both linguistic accuracy and geometric fidelity, thus improving modeling process explainability.

%% file: tables/mutimodal_comparison.tex
\setlength{\arrayrulewidth}{1pt}

\begin{table}[t]
\centering
\caption{Baseline Model Performance Comparison}
\label{tab:baseline_comparison}
\vspace{0.6em}

\begin{tabular*}{\columnwidth}{@{\extracolsep{\fill}}lcc}
\toprule
\textbf{Metric} & \textbf{Qwen2.5-7B} & \textbf{Qwen2.5-7B-VL} \\
\midrule

\textit{Exact Match Score} & 0.0058 & 0.0099 \\
\textit{BLEU Score} & 97.1827 & 97.9309 \\
\textit{Test Samples} & 3644 & 3644 \\
\textit{Command-Level F1} & 0.3247 & 0.3670 \\
\textit{Tolerance Accuracy} & 0.5016 & 0.4630 \\
\textit{Partial Match Rate} & 0.5554 & 0.6162 \\

\bottomrule
\end{tabular*}
\end{table}

%% file: sections/Conclusion.tex
\section{CONCLUSION}
In this study, we presented SldprtNet, a large-scale multimodal CAD dataset designed to advance research in semantic-driven 3D modeling, geometric deep learning, and cross-modal learning. Each of the 242,000+ samples includes a CAD model in .step and .sldprt formats, a multi-view composite image, a parameterized modeling script, and a description of the part’s appearance and function, capturing both precise geometry and design intent. By aligning these modalities, SldprtNet offers a rich training ground for models connecting language, vision, and CAD geometry \cite{Du2021NeuralShapeParser}.

To assess its effectiveness, we conducted baseline experiments comparing unimodal (Encoder\_text only) and multimodal (image + Encoder\_text) models using various evaluation metrics. Results clearly show that the multimodal dataset yields superior performance in CAD generation tasks.

We further augmented the dataset with custom encoder and decoder tools, enabling lossless conversion between CAD models and textual instructions. This supports fully parametric and interpretable text-to-CAD modeling: language models can output executable CAD operations, which can be converted back into text for verification. SldprtNet thus facilitates semantic-driven CAD generation, where high-level concepts in natural language are translated into precise 3D designs \cite{Jacomino2023Mistral7B}. We believe the dataset and toolchain will support future research in language-based CAD generation, cross-modal engineering understanding, and human-AI collaborative design. In future work, we aim to expand the dataset’s scale and feature coverage, and evaluate its utility in real-world text-to-CAD scenarios to further enable intelligent 3D design automation.